\documentclass[sigconf,nonacm]{acmart}

\usepackage{cleveref}
\usepackage{algorithm}
\usepackage{algorithmic}
\usepackage[most]{tcolorbox}
\usepackage{soul} 
\usepackage{mdframed}

\AtBeginDocument{%
  }

\setcopyright{acmlicensed}
\copyrightyear{}
\acmYear{}
\acmDOI{XXXXXXX.XXXXXXX}
\acmConference[]{}{}{}

\acmISBN{}

\begin{document}

\title{Unsupervised Urban Land Use Mapping with Street View\\ Contrastive Clustering and a Geographical Prior}

\author{Lin Che}
\email{linche@ethz.ch}

\affiliation{%
  \institution{ETH Zurich}
  \country{Switzerland}
}

\author{Yizi Chen}
\email{yizi.chen@ethz.ch}

\affiliation{%
  \institution{ETH Zurich}
  \country{Switzerland}
}

\author{Tanhua Jin}
\email{tanhua.jin@ugent.be}

\affiliation{%
  \institution{Ghent University}
  \country{Belgium}
}

\author{Martin Raubal}
\email{mraubal@ethz.ch}

\affiliation{%
  \institution{ETH Zurich}
  \country{Switzerland}
}

\author{Konrad Schindler}
\email{schindler@ethz.ch}

\affiliation{%
  \institution{ETH Zurich}
  \country{Switzerland}
}

\author{Peter Kiefer}
\email{pekiefer@ethz.ch}

\affiliation{%
  \institution{ETH Zurich}
  \country{Switzerland}
}

\renewcommand{\shortauthors}{Che et al.}

\begin{abstract}
Urban land use classification and mapping are critical for urban planning, resource management, and environmental monitoring. Existing remote sensing techniques often lack precision in complex urban environments due to the absence of ground-level details. Unlike aerial perspectives, street view images provide a ground-level view that captures more human and social activities relevant to land use in complex urban scenes. Existing street view-based methods primarily rely on supervised classification, which is challenged by the scarcity of high-quality labeled data and the difficulty of generalizing across diverse urban landscapes. This study introduces an unsupervised contrastive clustering model for street view images with a built-in geographical prior, to enhance clustering performance. When combined with a simple visual assignment of the clusters, our approach offers a flexible and customizable solution to land use mapping, tailored to the specific needs of urban planners. We experimentally show that our method can generate land use maps from geotagged street view image datasets of two cities. As our methodology relies on the universal spatial coherence of geospatial data (``Tobler's law''), it can be adapted to various settings where street view images are available, to enable scalable, unsupervised land use mapping and updating. The code will be available at \url{https://github.com/lin102/CCGP}.
\end{abstract}



\maketitle

\section{Introduction}
The rapid pace of urbanization has led to significant changes in urban landscapes. Urban land use maps are an indispensable tool for understanding and monitoring such transformations of the structure and function of cities \cite{hashem2015change, avtar2019population}. These maps depict the interactions between human activities and geographical space, highlighting how land is allocated for various socio-economic purposes, such as residential, commercial, and industrial uses \cite{herold2003spatial, jokar2013toward}. Monitoring and mapping urban land use has profound social impacts, by providing essential base data for urban planners, government officials, and environmental decision-makers in support of planning, governance, and sustainable development \cite{liu2017classifying}. Despite their importance, land use maps are incomplete or outdated in many cities \cite{kuemmerle2013challenges}.

\begin{figure}[t]
\centering
\includegraphics[width=0.95\columnwidth]{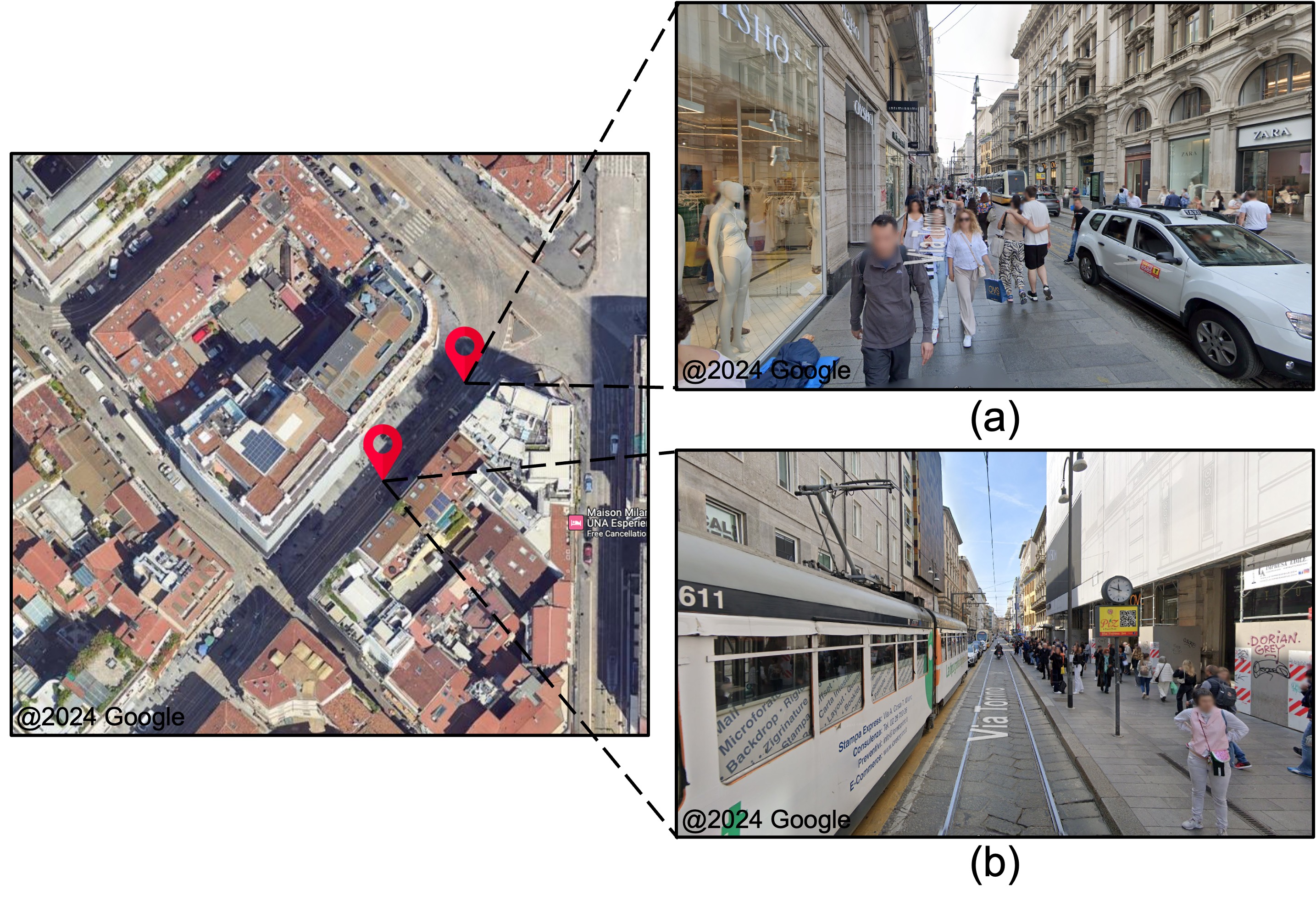} 
\caption{A motivational example illustrating the importance of spatial consistency. Two SVIs were taken within 20 meters of each other. Image (a) shows numerous pedestrians and shops, suggesting a commercial area. However, the adjacent image (b) contains dynamic objects such as trams and construction activities that occlude many of those cues, leading to significant visual differences between SVIs of the same land use type.}
\label{fig_motivation}
\end{figure}

Urban land use classification is especially challenging in densely populated megacities, where land use patterns are diverse and complex \cite{yin2021integrating}. With the increasing demand for efficient land use monitoring, rapid advances in machine learning and deep learning technology, and the growing availability of diverse data sources like remote sensing satellite imagery and street view images (SVI), significant progress has been made towards leveraging big data and machine learning for the automatic identification of urban land use \cite{zhang2019joint, campos2020understanding}. While high-resolution remote sensing images combined with computer vision techniques have proven effective for land cover classification \cite{tong2020land, mohanrajan2020survey}, challenges remain in land use classification. Unlike land cover, which reflects physical surface features, urban land use primarily concerns socio-economic attributes. Remote sensing and aerial images are limited by their top-down perspective, inherently lacking ground-level detail and land use-related visual information, e.g., on building facades \cite{cao2018integrating, qiao2021urban}. As shown in \Cref{fig_motivation}, it is challenging to identify specific land use types from birds-eye views, whereas ground-level street view images contain richer visual elements that make it easier to determine the urban function (in the example, a commercial area). Although many studies have used remote sensing images to predict land use classes \cite{huang2018urban}, the inherent limitations of this data source lead to low classification accuracy and hinder fine-grained analysis\cite{srivastava2020fine}. To overcome these limitations, recent studies have explored the use of ground-based images (e.g., Google Street View, Flickr) as alternative data sources that enable close-range perception and capture more detailed ground-level information \cite{li2017building, tracewski2017repurposing, wu2023mixed}. While street view images provide rich visual details, such as facade appearances and street furniture, they are inherently constrained by fixed camera viewpoints, occlusions from dynamic objects, and a limited ability to capture the broader spatial context. \Cref{fig_motivation} illustrates this with two SVIs taken within 20 meters of each other on the same road segment. Image (a) shows numerous pedestrians and shops, suggesting a commercial zone. However, in the adjacent image (b) dynamic objects like trams and construction scaffolds occlude many of those cues, greatly altering the visual appearance. Few studies have addressed the integration of spatial context \cite{fang2022spatial}. Another challenge is that existing SVI-based methods heavily rely on supervised training with labeled data \cite{fang2021synthesizing}, which is difficult to obtain in high quality. Moreover, visual differences between cities, due to social or environmental factors, mean that supervised models will not generalize well from one city to another, limiting their usability and impact.

To address these challenges, we introduce a self-supervised contrastive clustering model for panoramic street view images that exploits their spatial context. Based on this model, we demonstrate effective unsupervised land use clustering and mapping in two cities. Our primary contributions are twofold:
\begin{itemize}
\item We propose a contrastive clustering model with a geographical prior (CCGP). Our model learns to visually cluster images in an unsupervised fashion, in such a way that plausible geospatial neighborhood relations are preserved, accounting for the local smoothness of geographical patterns (``Tobler's law'').
\item Based on that end-to-end clustering model, we present the (to our knowledge) first unsupervised land use mapping scheme based on SVIs. Our CCGP-PCVA framework offers flexibility for urban planners and environmental decision-makers, enabling the creation of land use maps tailored to specific nomenclatures and professional needs. It does not require pre-existing labeled data, making it applicable to any city where there are SVIs.
\end{itemize}

\section{Related Work}
\subsection{Land Use Classification}
With the advancements in high-resolution remote sensing and deep learning, many land use classification methods have utilized satellite images \cite{talukdar2020land, digra2022land, li2024deep, wu2023mixed}. However, these methods face limitations due to the top-down perspective of remote sensing images, which lack detailed ground-level visual information related to land use \cite{huang2018urban}.

Street view images (SVI) have emerged as a valuable alternative data source, particularly for the classification of complex and heterogeneous urban land types \cite{zhu2015land, zhang2017parcel, cao2018urban}. SVIs are typically collected by map service
providers, such as Google Maps and Baidu Maps. They are captured along roadways using various collection methods, including vehicle-mounted and pedestrian equipment, and now cover most countries with rapid
ongoing expansion \cite{anguelov2010google}. \citet{srivastava2019understanding} employed machine learning algorithms to integrate Google Street View and OpenStreetMap annotations and predict land use at the object level. Furthermore, \citet{fang2021synthesizing} introduced the concept of location semantics, incorporating spatial relationships among SVIs, land parcels, and roads to enhance classification accuracy. However, prior studies often overlooked the spatial context beyond the visual modality of SVIs. The land use types of spatially close land parcels exhibit strong spatial correlation \cite{liu2016incorporating}, and classification should account for that. Relying solely on visual features from individual SVIs can result in geographically implausible discontinuities, as demonstrated in Figure \ref{fig_motivation}. We argue that suitable features for the analysis are those that are typically shared by spatially adjacent SVIs (like that in a commercial area, there are more shops and people on the streets than in an industrial one). A notable study by \citet{fang2022spatial} addressed the spatial context by designing a Spatial Context Graph Convolution Network (SC-GCN), transforming the land use classification task into a graph node labeling problem. 

There has been extensive research about unsupervised land cover classification in remotely sensed images \cite{duda2002unsupervised, campbell2011introduction, ghosh2022clustering}. For instance, the European Space Agency utilizes multispectral images to update land cover maps using unsupervised methods \cite{paris2019novel}. Prominent unsupervised classification techniques include $k$-means, Hierarchical Clustering, and ISODATA, among others \cite{jia2002cluster, richards2022clustering}. Current SVI-based studies predominantly rely on supervised classification methods. However, for land use classification tasks, ground truth labels for land use (e.g., from OSM) often suffer from widespread data gaps, spatial discontinuity, uneven distribution, and noise, with quality and quantity varying significantly across regions \cite{yeow2021point}. Furthermore, visual differences between cities hinder the model's ability to generalize effectively from one city to another \cite{zhu2019fine}.

In this paper, we enhance the clustering and spatial consistency of land use classification from SVIs by incorporating spatial proximity as a supervision signal, which also effectively mitigates the limitation of a single fixed viewpoint’s restricted field of view, leading to a more reasonable mapping. We propose the (to our knowledge) first self-supervised model based on SVIs, thus eliminating the need for labeled training data and addressing the challenge of transferring the model across cities.

\begin{figure*}[t]
\centering
\includegraphics[width=0.95\textwidth]{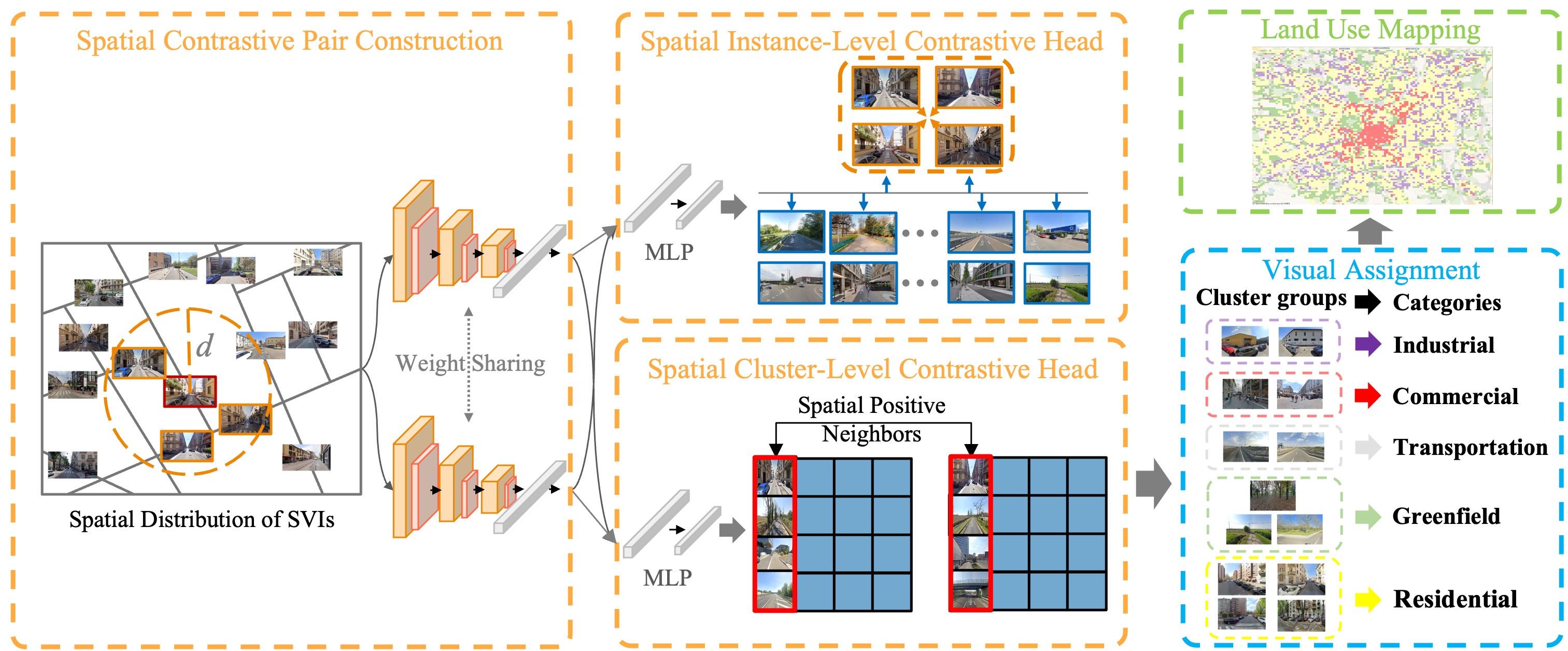} 
\caption{The proposed framework consists of three components: CCGP network (orange), PCVA (blue), and Grid Map Generation (green). CCGP selects spatially close images as positive samples, enriches them with data augmentation, and learns instance and cluster representations. PCVA assigns land use labels to clusters by manual interpretation of a representative high-confidence SVI for each cluster. Grid Map Generation aggregates the land use labels assigned to SVIs into a dense raster map.}
\label{fig_network}
\end{figure*}

\subsection{Contrastive Clustering}
As a self-supervised representation learning method, contrastive learning has recently achieved remarkable success in computer vision \cite{chen2020big, he2020momentum}. The core principle of contrastive learning is to map original data into a feature space where the distances between pairs of similar images ("positive pairs") are minimized, while those between dissimilar images ("negative pairs") are maximized. Positive pairs are typically generated through various data augmentation techniques applied to the same image. Several loss functions have been proposed for contrastive learning, including triplet loss \cite{schroff2015facenet} and InfoNCE loss \cite{oord2018representation}. \citet{khosla2020supervised} introduced the supervised contrastive (SupCon) loss, which clusters points of the same class in the embedding space while simultaneously pushing apart clusters of samples from different classes. In this paper, we adapt the SupCon loss to an unsupervised setting, leveraging a geographical prior instead of class labels to guide the learning process.

Deep clustering has proved to be more effective than traditional clustering methods like $k$-means \cite{macqueen1967some}, by leveraging the powerful feature extraction capabilities of deep learning. For instance, Invariant Information Clustering (IIC) leverages mutual information for label prediction in an end-to-end framework \cite{ji2019invariant}. Building on the powerful representational capabilities of contrastive learning, many studies have integrated clustering objectives into these frameworks, achieving good performance. E.g., PICA \cite{huang2020deep} optimizes clustering by maximizing the confidence of partitioning, thereby achieving the most semantically meaningful data separation. MICE \cite{tsai2020mice} is a robust probabilistic clustering approach that combines the discriminative power of contrastive learning with the semantic structures identified by a latent mixture model within a cohesive framework. CC \cite{li2021contrastive} was the first to reveal that the columns of a feature matrix could serve as cluster representations. By optimizing both instance-level and cluster-level contrastive losses, the model can simultaneously learn representations and cluster assignments in an end-to-end manner. GCC  \cite{zhong2021graph}  constructs a visual similarity graph, treating similar images as positive pairs, thereby elevating instance-level consistency to cluster-level consistency. However, in real-world urban settings, where SVIs often exhibit high visual similarity, mining positive pairs solely through visual features is challenging, and geographic smoothness cannot be ensured.

For our problem of geotagged SVIs, both similarity in the visual appearance space and consistent clustering in the geographic space are required. We propose a contrastive clustering method for geotagged SVIs that simultaneously considers visual and spatial proximity.

\section{Method}

\subsection{Contrastive Clustering with Geographical Prior}

Our end-to-end clustering network architecture adopts a three-component structure similar to previous contrastive clustering structures, such as CC \cite{li2021contrastive}, GCC \cite{zhong2021graph}, and CCES \cite{yin2023effective}. As depicted in Figure \ref{fig_network}, it includes a Spatial Contrastive Pair Construction (SCPC), a Spatial Instance-Level Contrastive Head (SICH), and a Spatial Cluster-Level Contrastive Head (SCCH). SICH and SCCH utilize a shared CNN pair construction backbone. Our method first constructs spatial neighbor positive pairs based on the geographic coordinates and spatial neighborhood of SVIs in SCPC, applies data augmentation, and then projects them into a latent feature space. Subsequently, in SICH and SCCH, instance-level and cluster-level contrastive learning are performed in the row and column spaces of the feature matrix, respectively, by minimizing the corresponding contrastive losses.

\subsubsection{Geographical Prior for Positive Neighbor Construction}

Inspired by Tobler's ``First Law of Geography'', which states that ``everything is related to everything else, but near things are more related than distant things'' \cite{tobler1970computer}, our approach incorporates \textbf{spatial consistency} and \textbf{smoothness} by introducing a geographical prior. This principle suggests that SVIs of spatially adjacent land parcels are more likely to exhibit the same land use. I.e., we construct cross-instance positive pairs, expected to have similar land use types, by pairing an SVI with the top-\( K \) spatial Nearest Neighbors (KNN) within a distance parameter \( d \), as illustrated by the orange circle in \Cref{fig_network} (left). Formally, let \( I \) denote a geotagged street view image dataset containing \( N \) unlabeled samples from \( M \) clusters. The geographic coordinates of the dataset are represented by \( \mathbf{X} = \{ \mathbf{x}_1, \mathbf{x}_2, \ldots, \mathbf{x}_N \} \), where \( \mathbf{x}_i \in \mathbb{R}^2 \) indicates the latitude and longitude of the \( i \)-th sample. To efficiently identify top-\( K \) spatial nearest neighbors for each sample, we construct a KD-Tree \cite{bentley1975multidimensional} based on \( \mathbf{X} \). For each sample \( \mathbf{x}_i \), its nearest neighbors are denoted as \( \mathcal{N}_K(\mathbf{x}_i) = \{\mathbf{x}_{i,1}, \mathbf{x}_{i,2}, \ldots, \mathbf{x}_{i,K}\} \), where \( \mathbf{x}_{i,j} \) is the \( j \)-th nearest neighbor of \( \mathbf{x}_i \). To facilitate spatial distance calculations, geographic coordinates in the WGS84 coordinate system (EPSG:4326) are transformed into the Web Mercator projection (EPSG:3857). This allows distances to be measured in meters, ensuring consistent spatial proximity metrics. Let \( T \) denote a stochastic data augmentation function applied to each sample, where \( T \sim \mathcal{T} \) is sampled from a set of transformations \( \mathcal{T} = \{\text{Resized Crop} \allowbreak, \text{ColorJitter} \allowbreak, \text{Grayscale} \allowbreak, \text{Horizontal Flip} \allowbreak, \text{Gaussian Blur} \allowbreak\} \) \cite{chen2020simple, li2021contrastive}. Each image \( I_i \) is stochastically augmented by a transformation \( T \sim \mathcal{T} \), producing \( I'_i = T(I_i) \).

\subsubsection{Spatial Instance-Level Contrastive Head} 
We employ a two-layer MLP head to project the feature matrix into a subspace, aiming to maximize the similarity between spatially close positive samples while minimizing that of negative ones. Let  \( \mathbf{z}' = \{ \mathbf{z}'_1, \mathbf{z}'_2, \ldots \} \) denote the corresponding features in the mapped subspace of the augmented images \( I' = \{ I'_1, I'_2, \ldots \} \). We adapt the SupCon loss \cite{khosla2020supervised}, where instead of using ground truth class labels, we leverage the proximity prior and encourage spatially close points to share the same land use class. Consequently, our loss function is designed to minimize the distances among spatially positive neighbors and encourage the model to focus on shared features. We define the Spatial Instance-Level Contrastive Head (SICH) loss as follows:

\begin{equation}
\mathcal{L}_{\text{\textit{SICH}}} = \sum_{i \in I} -\frac{1}{|\mathcal{N}_K(i)|} \sum_{j \in \mathcal{N}_K(i)} \log \frac{\exp(\mathbf{z}'_i \cdot \mathbf{z}'_j / \tau_I)}{\sum_{a \in A(i)} \exp(\mathbf{z}'_i \cdot \mathbf{z}'_a / \tau_I)}.
\label{eq:sich_loss}
\end{equation}

Here, the symbol \( \cdot \) denotes the dot product, \( \tau_I \) is the instance-level temperature parameter. The index \( i \) denotes the anchor, \( j \in \mathcal{N}_K(i) \), where \( \mathcal{N}_K(i) \) represents the set of the spatial top \( K \) nearest neighbors of \( I_i \) . \( A(i) \equiv I \setminus \{i\} \) denotes the set of all samples except \( i \).

\subsubsection{Spatial Cluster-Level Contrastive Head}
Unlike the original Contrastive Clustering, which assumes that different augmentations of the same image should share the same cluster assignment \cite{li2021contrastive}, our method extends this principle by introducing a proximity prior. Specifically, we construct cross-instance positive pairs between proximal samples, based on the assumption that spatially close locations tend to belong to the same cluster. \( I' = \{I'_1, \ldots, I'_N\} \) represent augmentations of the original images. For each \( I_i' \), we randomly sample one spatial neighbor \( \tilde{I}'_j \) from \( \mathcal{N}_K(i) \), producing the augmented set \( \tilde{I}' = \{\tilde{I}'_1, \ldots, \tilde{I}'_N\} \). A two-layer MLP head projects the feature matrix into an \( M \)-dimensional space, where \( M \) is the number of clusters. Let \( \mathbf{q}'_i \) and \( \tilde{\mathbf{q}}'_j \) denote the projected cluster assignments of the anchor and the spatial neighbor, respectively. \( \mathbf{q}'_i \) and \( \tilde{\mathbf{q}}'_j \) represent which images in \( I'\) and \(\tilde{I}'\) be assigned to cluster \( i\). We define the loss as:

\begin{equation}
\mathcal{L}_{\text{\textit{SCCH}}} = -\frac{1}{M} \sum_{i=1}^{M} 
\log \left( 
\frac{\exp(\mathbf{q}'_i \cdot \tilde{\mathbf{q}}'_i / \tau_C)}{\sum_{j=1}^{M} \exp(\mathbf{q}'_i \cdot \tilde{\mathbf{q}}'_j / \tau_C)}
\right),
\label{eq:scch_loss}
\end{equation}
where \( \tau_C \) is the cluster-level temperature parameter.

Furthermore, to prevent trivial solutions, where all samples are assigned to the same cluster, we introduce a clustering regularization loss based on entropy, similar to SCAN \cite{van2020scan} and CC \cite{yin2023effective}. The regularization loss \( \mathcal{L}_E \) is defined as:

\begin{equation}
\mathcal{L}_E = \log(M) - \frac{1}{M} \sum_{l=1}^{M} \mathcal{Z}_l \log \mathcal{Z}_l,
\label{eq:reg_loss}
\end{equation}

where \( \mathcal{Z}_l \) represents the normalized cluster probability distribution across the dataset, defined as:

\[
\mathcal{Z}_l = \frac{\sum_{j=1}^N q_{lj}}{\sum_{i=1}^M \sum_{j=1}^N q_{ij}},
\]

and \( \mathbf{q} \in \mathbb{R}^{M \times N} \) is the cluster assignment probability matrix, where \( q_{lj} \) denotes the probability of sample \( j \) belonging to the cluster \( l \). By maximizing entropy, this regularization encourages balanced cluster assignments while avoiding mode collapse. Given that real-world land use distributions are often imbalanced, the weight of \( \mathcal{L}_E \) (\( \eta \)) can be set relatively low in practice to account for inherent data imbalance without compromising regularization effects.

The overall objective function is thus formulated as:

\begin{equation}
\mathcal{L} = \mathcal{L}_{\text{SICH}} + \lambda \mathcal{L}_{\text{SCCH}} + \eta \mathcal{L}_E,
\label{eq:total_loss}
\end{equation}

where \( \lambda \) and \( \eta \) are hyperparameters used to balance the contributions of the spatial instance-level contrastive loss (\( \mathcal{L}_{\text{SICH}} \)), the spatial clustering contrastive loss (\( \mathcal{L}_{\text{SCCH}} \)), and the entropy-based clustering regularization loss (\( \mathcal{L}_E \)). The model's training process is summarized in \Cref{alg:CCGP}.

\begin{algorithm}[t]
\caption{CCGP Algorithm}
\label{alg:CCGP}
\begin{algorithmic}
    \STATE \textbf{Input:} SVI dataset \( I = \{I_1, I_2, \dots, I_n\} \) with geolocation \( X = \{x_1, x_2, \dots, x_n\} \), where \( I_i \) is an image and \( x_i \in \mathbb{R}^2 \) is its geolocation. Number of clusters \( M \), spatial neighborhood distance \( d \). Number of spatial positive samples \( K \), number of training epochs \( N_{ep} \). Hyperparameters \( \lambda, \eta \) for balancing loss terms. Stochastic augmentations set \( \mathcal{T} \).
    \STATE \textbf{Output:} Cluster assignments for dataset \( I \) and trained model parameters \( \theta \).
    
    \STATE Initialize model parameters \( \theta \);
    \STATE Construct a KD-Tree for \( X \) to identify the top \( K \) nearest neighbors \( N_K(x_i) = \{x_{i1}, x_{i2}, \dots, x_{iK}\} \) for each sample \( x_i \). Cache \( N_K(x_i) \) for all \( x_i \in X \).
    \FOR{each epoch \( e = 1 \) to \( N_{ep} \)}
        \STATE \quad \textbf{1: } Sample a random mini-batch of anchor images.
        \STATE For each anchor image , construct positive samples from its spatial neighbors based on \( N_K(x_i) \).
        \STATE \quad \textbf{2: }  Apply stochastic augmentations \( T \sim \mathcal{T} \) to both anchors and their positive neighbors, generating augmented views.
        \STATE \quad \textbf{3: }  Compute \( \mathcal{L}_{\text{SICH}} \) loss using \Cref{eq:sich_loss}.
        \STATE \quad \textbf{4: }  Compute \( \mathcal{L}_{\text{SCCH}} \) loss using \Cref{eq:scch_loss}.
        \STATE \quad \textbf{5: }  Compute \( \mathcal{L}_E \) loss using \Cref{eq:reg_loss}.
        \STATE \quad \textbf{6: }  Compute total loss: \( \mathcal{L} \) using \Cref{eq:total_loss}.
        \STATE \quad \textbf{7: }  Update \( \theta \) using Adam to minimize \( \mathcal{L} \).
    \ENDFOR
\end{algorithmic}
\end{algorithm}

\subsection{Post-Clustering Visual Assignment (PCVA) for Land Use Mapping}
While unsupervised methods do not require training labels, a challenge is that the clusters do not directly denote semantic land use categories \cite{dike2018unsupervised,ezugwu2022comprehensive}. To address this, we draw inspiration from the unsupervised land cover classification of remote sensing images, where clustering is always followed by category assignment \cite{rozenstein2011comparison}. However, unlike remote sensing images, which allow simultaneous visualization of all clusters and spatial distributions, SVIs do not support such direct map-based interpretation. Therefore, we employ a post-clustering visual assignment method (PCVA), consisting of three main steps:

\textbf{Step 1 -- Over-clustering:} Perform clustering using the CCGP, the same as other unsupervised clustering methods. In practical land use classification applications, the distribution of land use types in real geographic regions is often unknown and likely imbalanced. Setting the number of clusters $M$ a bit too high encourages over-segmentation into fine-grained clusters and thereby avoids missing class boundaries. 

\textbf{Step 2 -- Visual Assignment:} Like unsupervised clustering methods, the clustering results require manual assignment of categories. I.e, an expert visually interprets representative SVIs and assigns appropriate land use categories according to their specific needs. Notably, in cases of over-segmentation this may involve merging multiple clusters into a single category. Unsupervised methods are particularly useful when faced with a new environment. In such scenarios, SVIs are more suitable than remote sensing data, as they provide rich visual information that supports the identification of land use categories whose appearance was not defined a-priori.

\textbf{Step 3 -- Grid Map Generation:} Generate a map by discretizing the target area into a regular grid at the desired granularity (e.g., 50m $\times$ 50m, 100m $\times$ 100m). The land use type for each grid cell is determined as the one with the highest cumulative probability over all SVIs within the cell.

\begin{table}[t]
\centering
\begin{tabular}{lcc}
\toprule
\textbf{Land Use Type} & \textbf{Milan} & \textbf{San Francisco} \\
\midrule
Residential    & 402 & 425 \\
Greenfield     & 287 & 443 \\
Commercial     & 344 & 309 \\
Industrial     & 126 & 383 \\
Transportation & 118 & 224 \\
\midrule
\textbf{Total} & 1277 & 1784 \\
\bottomrule
\end{tabular}
\caption{Number of examples per class in the test sets}
\label{tab:evaluation_sets}
\end{table}

\begin{table*}[t]
\centering
\begin{tabular}{lcccccccc}
\toprule
\textbf{Dataset} & \multicolumn{4}{c}{\textbf{Milan}} & \multicolumn{4}{c}{\textbf{San Francisco}} \\
\cmidrule(lr){2-5} \cmidrule(lr){6-9}
\textbf{Metrics} & \textbf{NMI} & \textbf{ARI} & \textbf{ACC} & \textbf{mF1} & \textbf{NMI} & \textbf{ARI} & \textbf{ACC} & \textbf{mF1} \\
\midrule
$k$-means & 0.145 & 0.080 & 0.343 & 0.339 & 0.127 & 0.089 & 0.356 & 0.345 \\
SC & 0.125 & 0.092 & 0.357 & 0.153 & 0.074 & 0.054 & 0.383 & 0.160 \\
AC & 0.129 & 0.076 & 0.318 & 0.292 & 0.138 & 0.096 & 0.350 & 0.334 \\
AE & 0.182 & 0.112 & 0.377 & 0.390 & 0.142 & 0.114 & 0.396 & 0.374 \\
PICA & 0.332 & 0.291 & 0.521 & 0.463 & 0.285 & 0.269 & 0.518 & 0.429 \\
GCC & 0.392 & 0.337 & 0.602 & 0.487 & 0.389 & 0.327 & 0.554 & 0.475 \\
CC & 0.363 & 0.284 & 0.566 & 0.451 & 0.351 & 0.340 & 0.598 & 0.536 \\
\midrule
CCGP (Ours) & 0.541 & 0.480 & 0.706 & 0.631 & 0.481& 0.446& 0.668& 0.596 \\
CCGP-PCVA (Ours) & 0.548& 0.517 & 0.755 & 0.732 & 0.497& 0.516& 0.737& 0.702\\
\bottomrule
\end{tabular}
\caption{Clustering and classification performance of different methods.}
\label{tab:metric_results}
\end{table*}

\section{Experiments}
\subsection{Dataset Description}
We selected 10 km$\times$10 km central urban areas of Milan (Italy) and San Francisco (CA, U.S.), two representative cities from different continents with distinct urban characteristics, for our experiments. The SVI data was sourced from the Google Street View API \cite{GoogleStreetViewAPI}. Based on the OpenStreetMap (OSM) road network \cite{OpenStreetMap}, we ensured that there is one SVI sampling point for every 100m segment of roads. To avoid redundant samples due to complicated, intertwined street layouts, we applied spatial DBSCAN clustering \cite{ester1996density} as a preprocess to filter out redundant points less than 10 meters apart. After filtering we ended up with 30,994 SVIs in Milan, respectively 30,730 in San Francisco. To mitigate biases caused by the viewing angle, we use images with 360-degree horizontal and 90-degree vertical viewfields, composed of four 640$\times$640 images stitched together to form a 2560$\times$640 image. This approach avoids distortions commonly found in panoramic SVIs and focuses on the relevant visual information about the environment, as it 
removes a large portion of the irrelevant sky and ground pixels. 

In line with previous work \cite{wu2023mixed}, we obtained land use labels from OSM and regrouped their original, fine-grained categories into five classes: Residential, Greenfield, Transportation, Industrial, and Commercial. It should be noted that OSM land use labels are sparse, noisy, and imbalanced. The unreliable nature and limited availability of these labels are a primary motivation for developing an unsupervised method. Within the available labeled data, we reduced the proportion of dominant categories, such as Residential, to create a more balanced evaluation dataset. All
evaluation data were manually screened and filtered to ensure reliability. The final class distribution of the evaluation dataset is shown in \Cref{tab:evaluation_sets}.

\subsection{Evaluation Metrics}
To evaluate \emph{clustering} quality, we use two standard clustering metrics, Normalized Mutual Information (NMI) and Adjusted Rand Index (ARI). To measure the correctness of the resulting classification relative to ground truth we show overall Accuracy (ACC), as well as mean per-class F1-score (mF1) to account for class imbalance. The optimal alignment between predicted clusters and ground truth classes is determined using the Hungarian algorithm \cite{kuhn1955hungarian}. 

Furthermore, we calculate Moran's $I$ \cite{moran1950notes} to assess the spatial consistency and smoothness of the clustering and classification results. Higher values indicate greater spatial auto-correlation, following the definition

\begin{equation}
I = \frac{N}{W} \cdot \frac{\sum_{i=1}^{N} \sum_{j=1}^{N} w_{ij}(x_i - \bar{x})(x_j - \bar{x})}{\sum_{i=1}^{N} (x_i - \bar{x})^2}\;,
\end{equation}
where $N$ is the total number of observations, $W$ is the sum of all weights $w_{ij}$, $w_{ij}$ is the spatial weight between observations $i$ and $j$, $x_i$ and $x_j$ are the observed values at locations $i$ and $j$, respectively, and $\bar{x}$ is the mean of the observed values. In this study, the spatial weights $w_{ij}$ are constructed as the inverse of the distance, $\frac{1}{d}$, where $d$ is the distance between observations $i$ and $j$. The threshold distance to define the neighborhood is set to 100 meters.

For categorical land use with \( M \) classes, the predicted values \( x_i \) are binarized for each class. The overall Moran's \( I \) is computed as a weighted average across the \( M \) classes,

\begin{equation}
I_{\text{wt}} = \sum_{m=1}^{M} \frac{n_m}{N} I_m
\end{equation}

\noindent where \( n_m \) is the number of observations for class \( m \), \( N \) is the total number, and \( I_m \) is the Moran's \( I \) calculated for class \( m \) using the binarized data. This weighted Moran's I provides a single measure of spatial auto-correlation that accounts for the ground truth class distribution. Reaching higher \( I \)-values provides evidence for the validity of our hypothesis that incorporating geographic proximity as a self-supervision signal improves spatial consistency.

\begin{figure*}[t]
\centering
\includegraphics[width=0.98\textwidth]{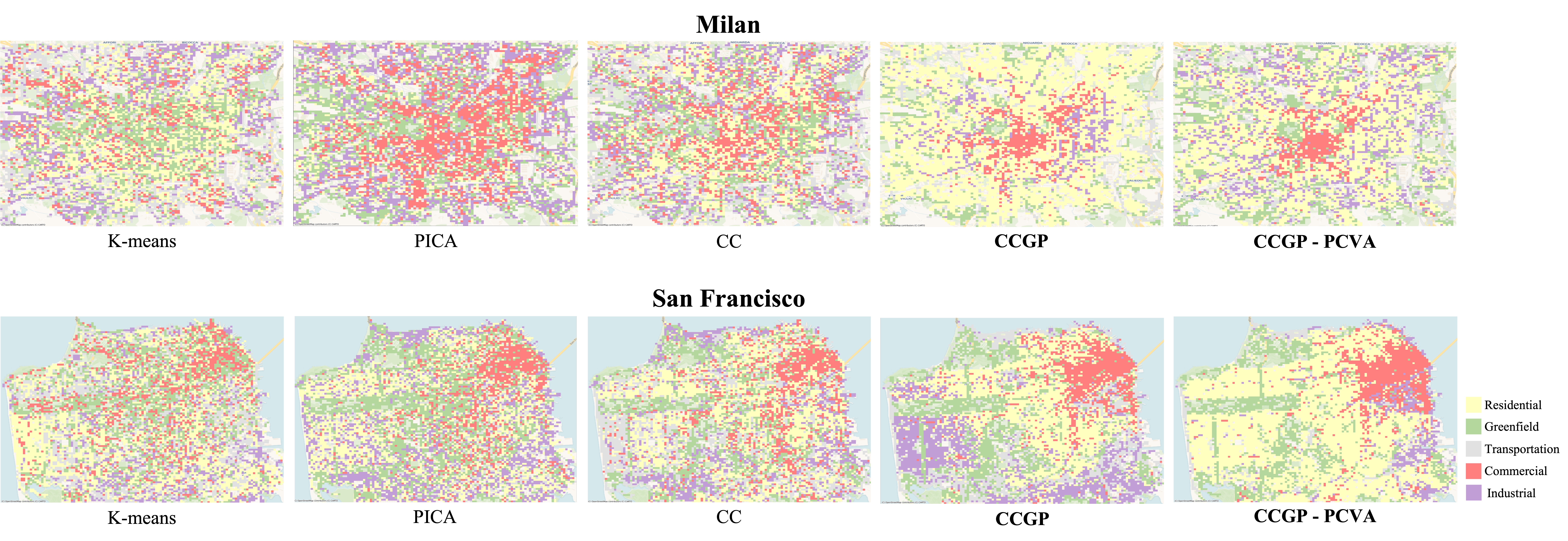} 
\caption{Land use mapping by clustering SVIs. Results at 100m$\times$100m grid resolution are shown for $k$-means, PICA, CC, CCGP, and CCGP-PCVA. Colors denote different land use categories as per the legend.}
\label{fig_mapping}
\end{figure*}

\begin{figure*}[t]
\centering
\includegraphics[width=0.8\textwidth]{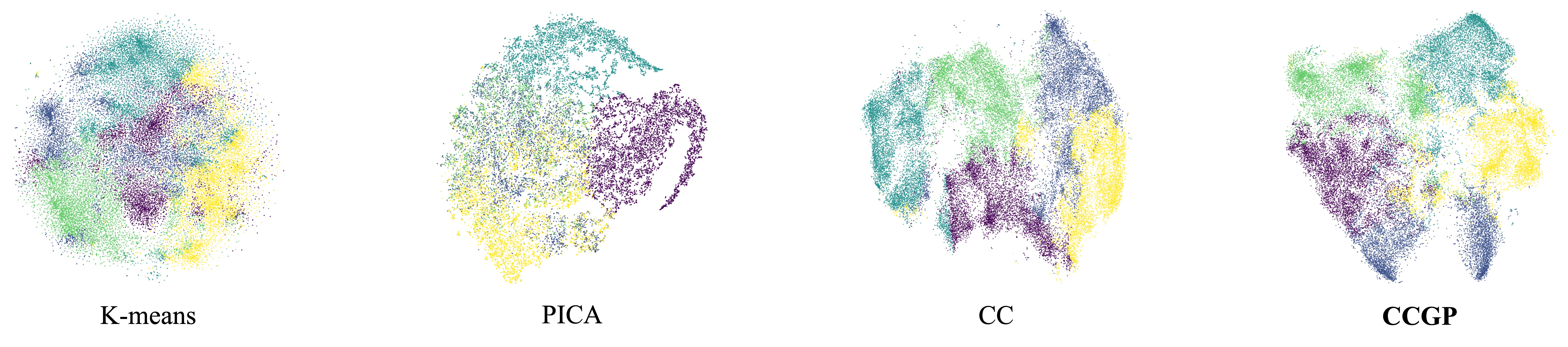}

\caption{{2D t-SNE visualization of features learned by $k$-means, PICA, CC, and CCCP for the San Francisco dataset.}}
    
\label{fig_tsne}
\end{figure*}

\subsection{Baseline Methods}

We evaluate the proposed CCGP against traditional and state-of-the-art clustering approaches, including $k$-means \cite{macqueen1967some}, Spectral Clustering (SC) \cite{ng2001spectral}, Agglomerative Clustering (AC) \cite{gowda1978agglomerative}, Autoencoder (AE)  \cite{bengio2006greedy}, partition confidence maximisation (PICA)  \cite{huang2020deep}, Contrastive Clustering (CC)  \cite{li2021contrastive}, and Graph Contrastive Clustering (GCC)  \cite{zhong2021graph}.

\subsection{Implementation Details}

All experiments were implemented using PyTorch \cite{paszke2017automatic}. In our framework, we employ ResNet-18 \cite{he2016deep} as the primary backbone and train the networks on a single Nvidia RTX 4090 GPU. We cluster into $m=5$ clusters and use 128-dimensional feature vectors $\mathbf{z}$ in the SIC head, with the spatial neighborhood set to $K=1$. The loss is minimized with the Adam optimizer, with an initial learning rate of 0.0002 and no weight decay. The batch size was set to 128, the instance-level temperature $\tau_I$ to 0.5, and the cluster-level temperature $\tau_C$ to 1.0. As weight parameters we set $\lambda=2$ and  $\eta=0.2$. The model was trained from scratch until convergence, which takes approximately 40 epochs.

\begin{table}[t]
\centering
\begin{tabular}{lcc}
\toprule
\textbf{Method} & \textbf{Milan} & \textbf{San Francisco} \\
\midrule
CC & 0.340 & 0.311 \\
CCGP & 0.519 & 0.448 \\
\bottomrule
\end{tabular}
\caption{Comparison of Moran's $I$ for basic CC and for our version based on spatial nearest neighbors}
\label{tab:weighted_moran}
\end{table}

\subsection{Experimental Results}
As shown in \Cref{tab:metric_results}, the CCGP framework outperforms the baselines by significant margins, across all metrics. The spatial auto-correlation of the result, measured via the weighted mean of Moran’s $I$ values, further confirms that the proposed geographic prior enhances spatial consistency of the clustering, \Cref{tab:weighted_moran}. The maps in \Cref{fig_mapping} qualitatively depict the effect. We also performed t-SNE \cite{van2008visualizing} on the features learned by different clustering schemes. As can be seen in \Cref{fig_tsne}, CCGP gives rise to particularly compact and well-separated clusters.

\subsection{Impact of Nearest Neighbor Count $K$}

\begin{figure}[t]
\centering
\includegraphics[width=0.7\columnwidth]{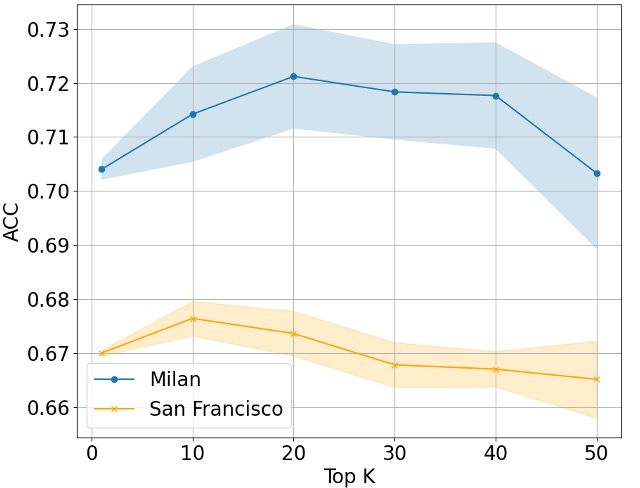} 
\caption{Mean accuracy for varying $K$ between 1 and 50. Shaded regions denote standard error bars.}
\label{fig_k_analysis}
\end{figure}

We conduct additional experiments to understand the impact of the Nearest Neighbor parameter \( K \) on model performance. As shown in \Cref{fig_k_analysis}, we run five training and testing runs with different random seeds for each value of \( K \). The results indicate that the model is most stable when \( K = 1 \), where one always selects the spatially closest sample to form a positive pair.
 As \( K \) increases, model performance goes up, at the cost of decreasing stability (higher fluctuation between different runs). Intuitively this makes sense: sampling more distant positive pairs is more informative as long as the distance does not exceed the true (average) auto-correlation length of the land use distribution, but increasing the distance also increases the likelihood of false positive samples. As \( K \) increases the model also takes longer to converge, likely as a consequence of the larger portion of false positives.

\begin{figure*}[t]
\centering
\includegraphics[width=0.85\textwidth]{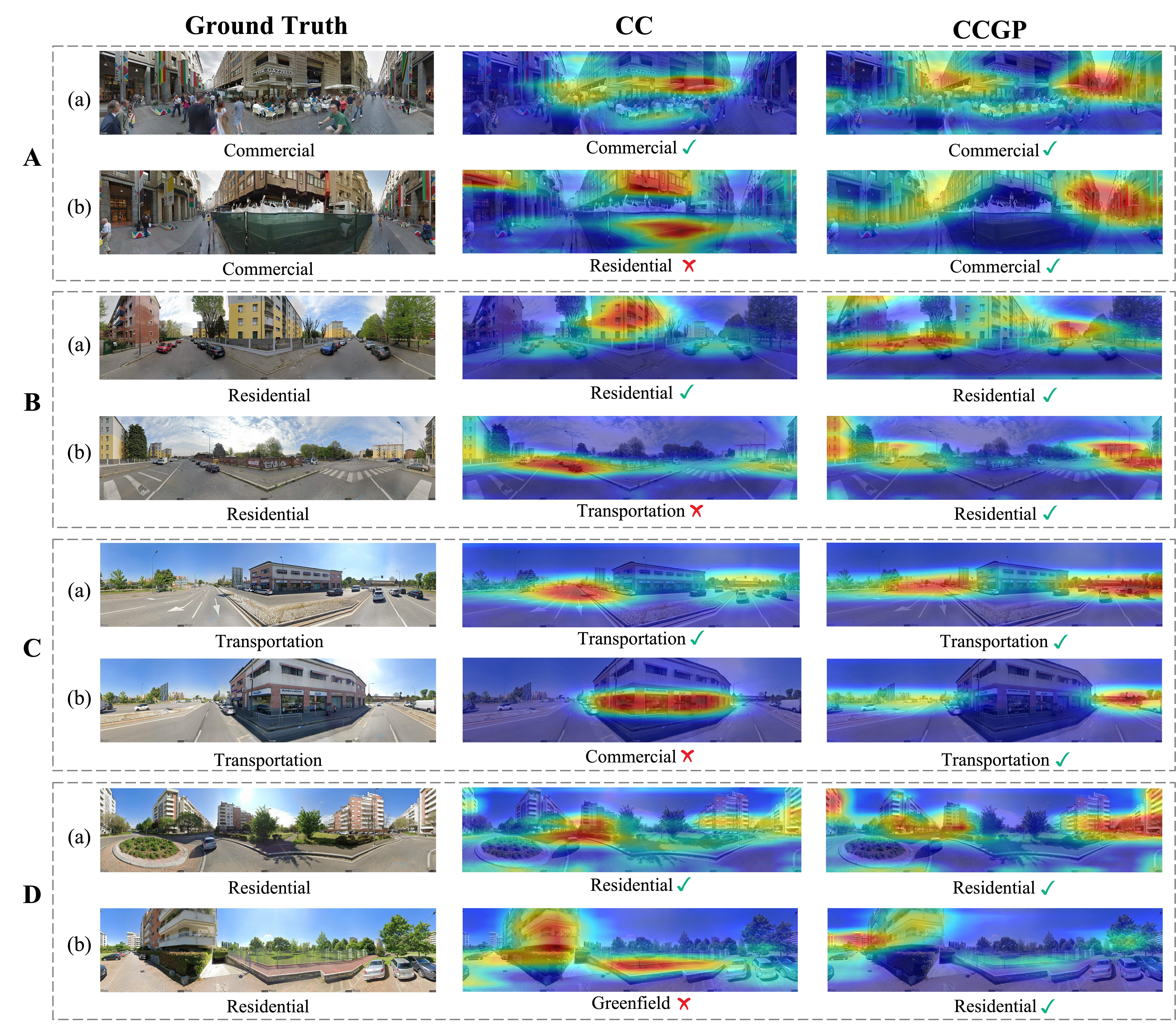} 
\caption{GradCAM++ heatmaps indicate the influence of different image regions on the cluster assignment. We show four sets of results (A–D). In each set, samples (a) and (b) are spatially adjacent. The left column displays SVIs with their ground truth labels, while the middle and right columns show the results and predictions of CC and CCGP, respectively.}
\label{fig_attention_map}
\end{figure*}

\subsection{Visualizing the Impact of the Geographical Prior}

We explore how the proposed, geographically modulated construction of positive pairs in the CCGP architecture alters the behaviour of the CC model. To illustrate this, we randomly selected pairs of spatially adjacent SVI samples where CCGP returned correct predictions, whereas CC did not. We utilize GradCAM++ \cite{chattopadhay2018grad, jacobgilpytorchcam} to trace which features, and consequently which image regions, have most influenced the cluster assignment. 

\Cref{fig_attention_map} showcases four sets of GradCAM++ heatmaps (A–D). In each set, (a) and (b) are spatially adjacent SVI samples. The left column displays the SVIs with their ground truth labels, the middle column shows the CC predictions along with their feature importance heatmaps, and the right column shows the corresponding CCGP results. In Set A, CC(b) fails to make a correct prediction due to a salient occlusion caused by temporary construction activities. In contrast, CCGP appears to rely more on features shared with adjacent SVIs, which are more indicative of the land use category. In Set B, the sample (b) is located at an intersection in such a way that category-specific visual cues, like residential buildings, occupy only a small portion of the viewfield. As a result, CC is misled to place the sample in the transportation cluster. CCGP again benefits from features shared among adjacent samples and relies on the small, yet essential residential buildings. Set C presents two consecutive sample points along a national highway. In image (b), the field of view is dominated by an auto service center on the roadside, impacting the CC result. In contrast, CCGP concentrates on features shared between adjacent transportation locations. Set D further demonstrates that contrastive learning based on spatially close locations enables the model to focus on features that are shared across changing viewpoints, which tend to be a better indication of their shared land use category.
In contrast, CC is driven entirely by features that are salient in the specific image in question, which is natural due to the contrastive learning from (an augmented version of ) single images. As a result, distractors that are prominent due to the individual viewpoint can more easily overwhelm the cues for the actual land use category. By learning to rely on cues that nearby viewpoints have in common, CCGP regularizes the feature representation in a way that favors visual elements shared within a reasonable geographic context, thus respecting spatial autocorrelation and leading to more stable and spatially consistent predictions.

\begin{figure}[t]
\centering
\includegraphics[width=0.95\columnwidth]{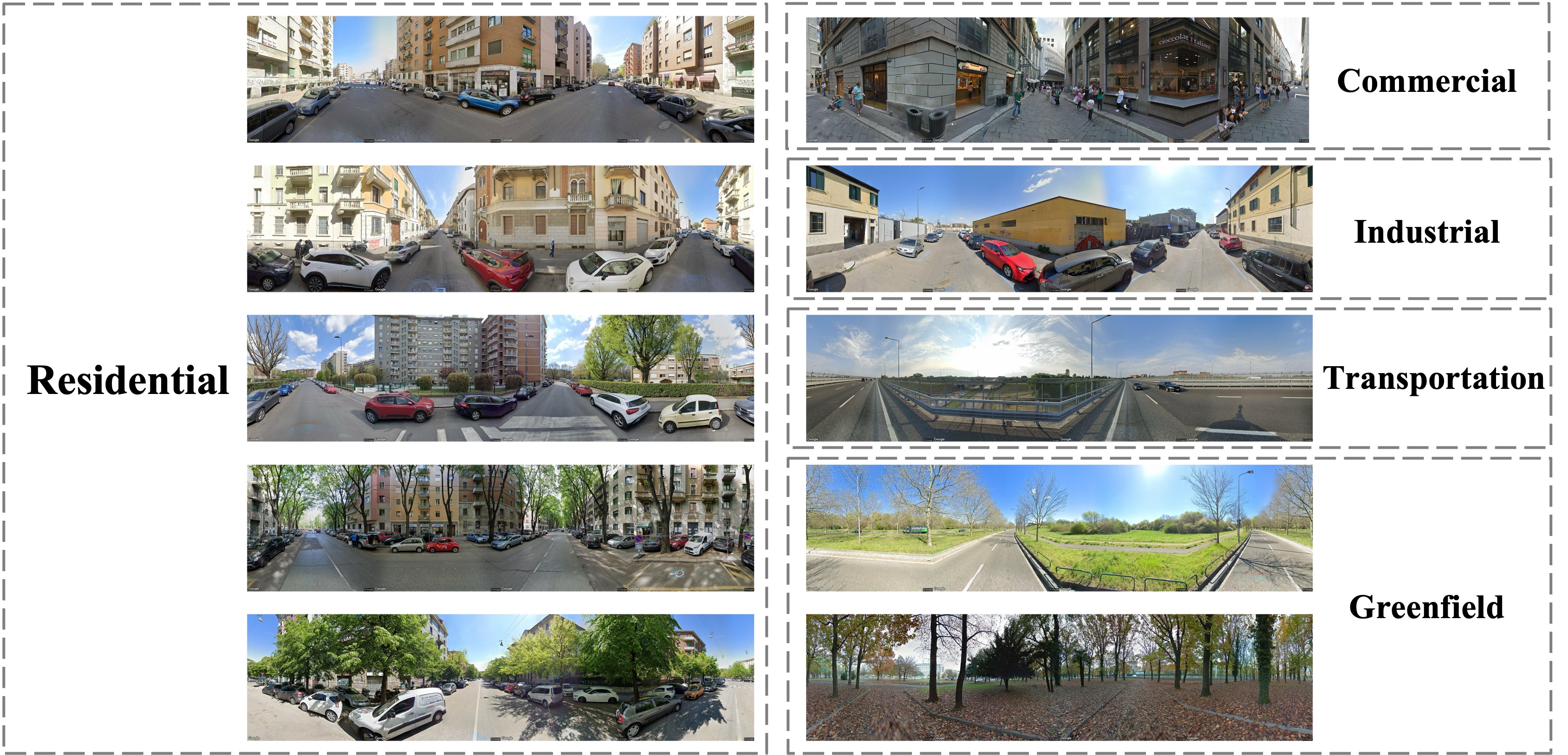} 
\caption{Results for Milan after Post-Clustering Visual Assignment. The ten images are the highest confidence samples from the over-clustering process. The grouping into five land use categories as shown is based on the visual assignment of these images.}
\label{fig_visual_assignment}
\end{figure}

\subsection{PCVA for Land Use Mapping}

Using Milan as an example, we further demonstrate our fine-grained post-clustering visual assignment (PCVA) method for land use mapping based on CCGP. The process involves three key steps:

\textbf{Step 1: Over-Clustering} -- In the above clustering experiments we had set the number of clusters to \( M = 5 \) for CCGP, for a meaningful evaluation in terms of the baselines and the true labels. To produce more fine-grained mapping results, we follow standard practice in unsupervised classification and increase the number of clusters to \( M = 10 \) for CCGP-PCVA. Using the trained clustering model, we output the representative images with the highest confidence for each class, as shown in \Cref{fig_visual_assignment}. 

\textbf{Step 2: Visual Assignment} -- Any clustering method, including CCGP as well as the baselines, requires a post-assignment of clusters to semantically defined classes. Based on the outputs of the previous step, the user can group and label the fine-grained clusters to associate them with land use categories in a many-to-one mapping that suits their needs. As shown in Figure \ref{fig_visual_assignment}, we categorize the five clusters on the left as residential, featuring various types of residential buildings. Despite differences in these environments -- e.g., some show more greenery while others capture denser inner city areas -- they are recognizable as residential. On the right, we interpret the prototypical images and label each cluster as commercial, industrial, transportation, or greenfield. The interpretation shown is our own, but the assignment of clusters to land use categories can be flexibly adjusted by the urban planner depending on their analysis or planning task. For instance, the broader "Residential" category could be subdivided into finer categories based on the density, or a separate "Forest" category could be split from "Greenfields". 

Compared to the visual interpretation of remote sensing images, SVIs provide richer visual information and require less location-specific prior knowledge and domain expertise. Moreover, we note that, beyond clustering, the representative SVIs of all clusters form a compact summary of the range of visual elements present in the target city, which we term the \emph{City Visual Profile} (CVP). We contend that, under the chosen clustering metric, these images can be seen as the most concise set that spans the range of visual features needed to characterize the target city. Borrowing the concept of the Minimal Description Length (MDL) in information theory~\cite{rissanen1978modeling}, which seeks the simplest possible model for a dataset by minimizing the code length needed to store it, one can think of the CVP as an approximation of the "Minimal Visual Description Length" for the target environment. We hypothesize that it may be useful also in other fields of research, such as urban perception studies or the development of social sensing techniques \cite{ito2024understanding}.

\textbf{Step 3: Grid Map Generation} -- To obtain a dense map, we resample the land use categories of all SVIs onto a regular grid with 100m $\times$ 100m resolution, assigning each grid cell the category with the highest cumulative probability over all images that fall into it. \Cref{fig_mapping} qualitatively compares the land use maps obtained for Milan and San Francisco with four different methods: $K$-means, CC, 5-cluster CCGP, and 10-cluster CCGP with subsequent PCVA. 

\subsection{Results and Discussion}
Looking at the results in \Cref{tab:metric_results} and \Cref{fig_mapping}, CC achieves more plausible clustering results and more realistic spatial land use patterns than $K$-means, presumably due to its superior representation learning capability. However, a significant amount of entropy and implausible spatial jittering remains. In contrast to the original CC, the spatial context that our CCGP approach injects into the contrastive clustering scheme greatly enhances spatial coherence and improves mapping performance. The heatmap analysis with GradCAM++ further reveals that CCGP regularizes the feature extraction to focus on visual elements that are shared across neighboring SVIs, and thus more indicative of the environment. As a consequence, it mitigates the impact of salient distractors and boosts the influence of subdued, but informative visual cues. Despite these improvements, CCGP has certain limitations. Real-world datasets often have imbalanced class
distributions.\footnote{This is also the reason why the mF1 metric is arguably preferable to the overall ACCuracy.} The imbalance means that rare categories are either underrepresented in terms of their visual appearance; or they are overrepresented in terms of their frequency, if counter measures are taken such as rebalancing the dataset. In future work, we aim to tackle this limitation by separating the visual characteristics from the frequency distribution and learning both from data. In the fine-grained CCGP-PCVA method, the larger number of clusters and the interactive post-assignment to some degree mitigate the imbalance problem. Consequently, CCGP-PCVA consistently delivers the best results.

\section{Conclusion}
We have proposed the first unsupervised approach to urban land use clustering and mapping based on street view images. Our Contrastive Clustering with Geographical Priors (CCGP) framework enhances clustering performance and ensures greater spatial coherence of the resulting maps. When combined with Post-Clustering Visual Assignment (PCVA) our method allows for a flexible and fine-grained categorization of urban land use, a property it inherits from the classical idea of clustering-based ``unsupervised classification''. Experiments on two real-world urban SVI datasets demonstrate the potential of our method to support urban land use classification at scale, across varying geographical contexts.

\begin{acks}
This work has received funding from the Swiss State Secretariat for Education, Research and Innovation (SERI). The research has been performed in the scope of Eyes4ICU, a project funded by the European Union under the Horizon Europe Marie Skłodowska-Curie Actions, GA No. 101072410. 
\end{acks}

\bibliographystyle{ACM-Reference-Format}
\bibliography{refs}

\end{document}